\documentclass[letterpaper]{article}
\usepackage{uai2020}
\usepackage[margin=1in]{geometry}
\usepackage{times}
\usepackage[round]{natbib} 
\usepackage{graphicx} 
\usepackage{hyperref}
\usepackage{footnote}
\usepackage{amsmath}
\usepackage{amssymb}
\usepackage{xargs} 
\usepackage[utf8]{inputenc}
\usepackage[disable]{todonotes}
\usepackage{subfig}
\usepackage{footmisc}
\usepackage{float}
\usepackage{booktabs}
\usepackage{xfrac}
\usepackage{cleveref}
\usepackage{soul}

\crefname{section}{\S}{\S\S}
\Crefname{section}{\S}{\S\S}
\crefname{table}{Tab.}{}
\crefname{figure}{Fig.}{}
\crefname{algorithm}{Algorithm}{}
\crefname{equation}{eq.}{}
\crefname{appendix}{App.}{}
\crefname{prop}{Proposition}{}
\crefformat{section}{\S#2#1#3}  

\newcommandx{\unsure}[2][1=]{\todo[linecolor=red,backgroundcolor=red!25,bordercolor=red,#1]{#2}}
\newcommandx{\change}[2][1=]{\todo[linecolor=blue,backgroundcolor=blue!25,bordercolor=blue,#1]{#2}}
\newcommandx{\info}[2][1=]{\todo[linecolor=OliveGreen,backgroundcolor=OliveGreen!25,bordercolor=OliveGreen,#1]{#2}}
\newcommandx{\improvement}[2][1=]{\todo[linecolor=yellow,backgroundcolor=yellow!25,bordercolor=yellow,#1]{#2}}
\newcommandx{\thiswillnotshow}[2][1=]{\todo[disable,#1]{#2}}

\usepackage{url}
\usepackage[safe]{tipa}

\newcommand{\method}{TX-Ray} 

\newcommand{\supervisedcorpus}{IMDB}

\newcommand{\interpretability}{interpretability}

\newcommand{\persample}{per-instance}

\newcommand{\unsupervisedcorpus}{WikiText-2}
\newcommand{\activationhistorgrams}{feature preference distributions} 
\newcommand{\activationhistorgram}{feature preference distribution}
\newcommand{\unsupervised}{unsupervised} 

\newcommand{\supimdbsuff}{imdb-sup}
\newcommand{\pt}{pre-training}
\newcommand{\explainability}{explainability}

\newcommand{\colbox}[2]{{\setlength{\fboxsep}{.14pt}\colorbox{#1}{\strut #2}}}
\let\vec\mathbf
\title{\method: Quantifying and Explaining Model-Knowledge Transfer in (Un-)Supervised NLP}
\author{Nils Rethmeier\thanks{Work done as part of the ``PhD School of Science'' program at Copenhagen University.}\\ 
        SLT Lab \\
        DFKI \\ 
        Berlin, Germany\\
        \texttt{nils.rethmeier@dfki.de}
        \And Vageesh Kumar Saxena\\
        SLT Lab \\
        DFKI \\ 
        Berlin, Germany\\
        \texttt{vageesh\_kumar.saxena@dfki.de}
        \And Isabelle Augenstein\\
        Department of Computer Science\\
        University of Copenhagen\\
        Denmark\\
        \texttt{augenstein@di.ku.dk}
        }

\author{Nils Rethmeier$^{* +}$ \text{ }  Vageesh Kumar Saxena$^{+}$ \text{ } 
\text{ } Isabelle Augenstein$^{*}$ \\
${}^{+}$Speech and Language Technology Lab, DFKI, Berlin, Germany \\
${}^{*}$Department of Computer Science, University of Copenhagen, Denmark \\
{\tt \{nils.rethmeier,vageesh\_kumar.saxena\}@dfki.de, augenstein@di.ku.dk}
}
\date{}
\begin{document}
\definecolor{unsup_imdb_color}{HTML}{B6D7A8}
\definecolor{unsup_imdb_dark_color}{HTML}{62c936}
\definecolor{unsup_wiki_color}{HTML}{e7a5a5}
\definecolor{unsup_48_wiki_color}{HTML}{ebc3c3}
\definecolor{unsup_1_wiki_color}{HTML}{949494}
\definecolor{sup_imdb_color}{HTML}{A4C2F4}
\definecolor{sup_imdb_ltsm_mx_color}{HTML}{B4A7D6}
\definecolor{length_reduced}{HTML}{E06566}
\definecolor{length_increased}{HTML}{6D9EEB}

\definecolor{t}{HTML}{e06666}
\definecolor{r1}{HTML}{D7737A}
\definecolor{a}{HTML}{CE808E}
\definecolor{n}{HTML}{C68DA2}
\definecolor{s}{HTML}{BD9AB7}
\definecolor{f}{HTML}{B5A7CB}
\definecolor{e}{HTML}{ACB4DF}
\definecolor{r2}{HTML}{A4C2F4}
\newcommand{\transfer}{{\setlength{\fboxsep}{.14pt}\colbox{t}{t}\colbox{r1}{r}\colbox{a}{a}\colbox{n}{n}\colbox{s}{s}\colbox{f}{f}\colbox{e}{e}\colbox{r2}{r}{\strut }}}

\maketitle
\begin{abstract}
While state-of-the-art NLP explainability (XAI) methods focus on explaining \emph{per-sample} decisions in \emph{supervised} end or probing tasks, this is insufficient to explain and quantify \emph{model knowledge transfer} during (un-)supervised training.
Thus, for TX-Ray, we modify the established computer vision \explainability\ principle of `visualizing preferred inputs of neurons' to make it usable for both \emph{NLP} and for \emph{transfer analysis}. This allows one to analyze, \emph{track and quantify} how \emph{self- or supervised} NLP models first \emph{build} knowledge abstractions in pretraining (1), and then \emph{transfer} abstractions to a new domain (2), or \emph{adapt} them during supervised fine tuning (3) -- see \cref{fig:model}. TX-Ray expresses neurons as \activationhistorgrams\ to \emph{quantify fine-grained knowledge transfer or adaptation} and \emph{guide human analysis}. We find that, similar to Lottery Ticket based pruning, TX-Ray based pruning can \emph{improve test set generalization} and that it can reveal how early stages of self-supervision automatically learn linguistic abstractions like parts-of-speech.
\end{abstract}

\begin{figure}[t]
    \centering
    \includegraphics[width=\linewidth]{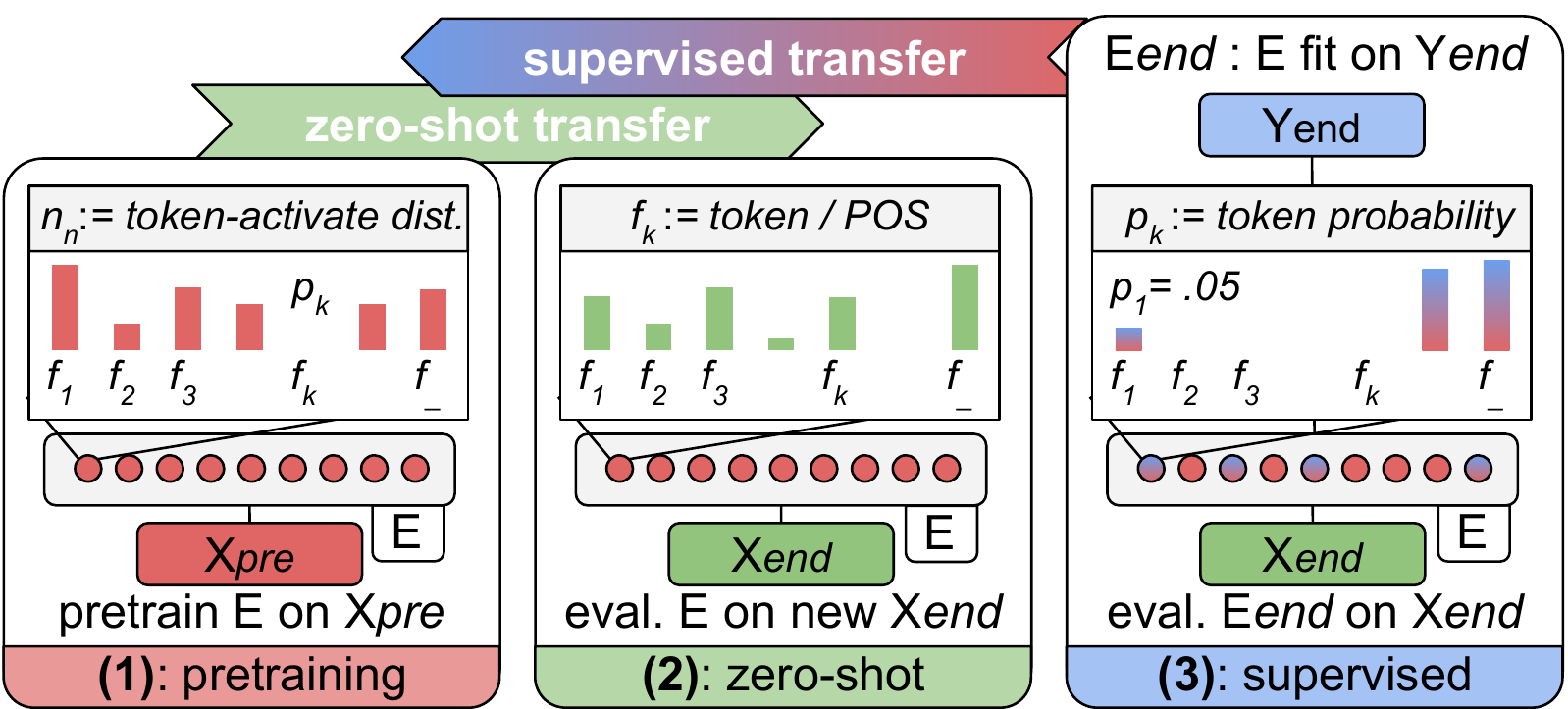}
    \caption{\textbf{Example uses of \method:} for transfer learning and model \interpretability. \textbf{Left (1):} \colbox{unsup_wiki_color}{pre-train} a sequence encoder $E$ on corpus $X_{pre}$ and collect \activationhistorgrams\ (\cref{sec:neurons_as_dist}, red bars) over input features (e.g. words) $f_k$.\protect\footnotemark\ \textbf{Middle (2):} apply, but not re-train, the encoder $E$ to \colbox{unsup_imdb_color}{new domain inputs} $X_{end}$ and observe the \emph{changed neuron activation} (green). Similarities in red and green reveal zero-shot forward transfer potential or data match between $X_{pre}$ and $X_{end}$ according to $E$. \textbf{Right (3):} fine-tune \colbox{unsup_wiki_color}{encoder} $E$ on supervision \colbox{sup_imdb_color}{labels} $Y_{end}$ to reveal `backward' \transfer\ of supervision knowledge into the encoder's knowledge abstractions.\protect\footnotemark}\label{fig:model}
\end{figure}
\addtocounter{footnote}{-2} 
\stepcounter{footnote}\footnotetext{Features $f_k$ are discrete inputs like tokens or POS tags.}
\stepcounter{footnote}\footnotetext{`Backward transfer' since $E$ changes, but labels $Y$ do not.} 

\section{INTRODUCTION}
Continual and Transfer Learning have gained importance across fields like NLP, where the de facto standard approach is to pretrain a sequence encoder and fine-tune it to a set of supervised end-tasks \citep{TuneNotTune19}. Analysis and understanding of transfer in NLP are currently focused on using either supervised probing tasks \citep{belinkov-glass-2019-analysis} to compare task performance metrics \citep{Glue} or laborious per-instance explainability \citep{belinkov-glass-2019-analysis}. \emph{Supervised} probing annotation is costly, but not guaranteed to be reliable under domain shifts. Probing is also limited to analyzing foreseen (probed) \emph{knowledge absorption} aspects, while unforeseen, model-knowledge properties that underlie and thus further our understanding of self-supervised pretraining remain hidden \citep{RightWrong19}. In fact, `decision understanding' explainability techniques, as \cite{CSI19} term them, compute the relevance of a feature or neuron for an end-task prediction score. This makes `decision understanding' explainability unable to answer the following research questions (\textbf{RQ1-3}) -- i.e. how can we explain transfer?

\textbf{(RQ1), \unsupervised\ knowledge absorption:} Can explainabilty (XAI) analyze how self-supervised models build and change knowledge abstractions during pretraining and can XAI measure knowledge changes? Do measures coincide with conventional metrics like perplexity? If and when does self-supervision learn linguistic abstractions like word function (parts-of-speech)? 

\textbf{(RQ2), zero-shot knowledge transfer:} What knowledge subset do pretrained models apply to a new domain without re-training, e.g. in a zero-shot setting?

\textbf{(RQ3), supervised/ backwards transfer:} Can knowledge transfer `backwards' from supervision labels into a pretrained model? Does XAI identify which neurons are reconfigured -- i.e.\ become task (ir)relevant due to supervision. Can we validate XAI-based transfer measures (RQ1) empirically by pruning (ir)relevant neurons? 

\textbf{TX-Ray can analyze and \emph{quantify} (self-)supervised model knowledge change:}
To answer RQ1-3 we propose TX-Ray. TX-Ray -- i.e., Transfer eXplainability as pReference of Activations analYsis -- modifies the well established activation maximization method of visualizing the preferred inputs of neurons \citep{ActivationMaximization} to suit NLP. The resulting fine-grained `model understanding' -- as \cite{CSI19} term it -- enables us to \emph{quantify knowledge changes or transfer} during training at the level of individual neurons -- without requiring or preemptively limiting analysis to probing task supervision semantics. The method is designed to explore model knowledge change at both neuron (detail) and model (overview) level to enable concise or deep explorative analysis of unforeseen knowledge transfer mechanics to help us better analyze (continual) transfer, model knowledge generalization \citep{RightWrong19,Lottery1}, or low-resource learning.  
\cite{XAISaliency, XAILies} showed that XAI methods do not guarantee faithful explanations. We thus use TX-Ray's transfer measures to guide neuron pruning and  empirically verify that it can identify task (ir)relevant neurons that boost or lower test set generalization as expected. We also demonstrate that supervision not only causes catastrophic forgetting of knowledge, but also adds new knowledge into previously un-preferred (under-used) neurons (\cref{tab:Pruning}).

\section{APPROACH}\label{sec:approach}
TX-Ray is inspired by the widely used activation maximization explainability method, which is based on the idea that ``a pattern to which a unit is responding maximally is a good first-order abstraction of what a unit (neuron) is doing. A simple way is to find the input samples that produce the highest activation for a neuron. Unfortunately, this opens the problem of how to `combine' these samples.'' \citep{ActivationMaximization}. In computer vision, naively combining image maximum feature activation maps ``over a corpus does not produce interpretable results'' \citep{ActivationMaximization}. In NLP, however, maximal activations of discrete token feature can easily be combined over many samples to form a discrete distribution of `tokens that a neuron prefers'. These corpus-wide input \activationhistorgrams\ let us visualize how each neuron abstracts input knowledge subsets.

A major advantage of a `feature preference' method is that it can analyze \emph{non-supervised models over an entire corpus}, while `prediction score relevance explainability' methods require \emph{supervised models}, and \emph{only explain individual instances} \citep{belinkov-glass-2019-analysis}. When representing a neuron's abstracted knowledge as a \activationhistorgrams, we can measure knowledge change, or transfer, during learning using standard measures such as Hellinger Distance -- i.e., a symmetric version of the Kullback Leibler divergence. This allows one to track changes in neuron knowledge abstractions during model pretraining, model application to new domains or due to supervised fine tuning -- see experimental section. Additionally, we automatically determine neurons that change their knowledge the most over time to provide interesting starting points (see \cref{fig:ZS_hellinger_length}, \ref{fig:Supervised_hellinger_length}) for nuanced, per-neuron analysis (see \cref{fig:ZS637} and \ref{fig:supervised_877}).  

\subsection{NEURONS AS FEATURE PREFERENCE}\label{sec:neurons_as_dist}
We thus expresses each neuron $n_n$ as a distribution over preferred features $f_k$ with activation probabilities $p_k$ (\cref{fig:model}) that have been aggregated over an entire corpus to construct each $n_n$ distribution as follows. 

\textbf{(1) Record what features neurons prefer:}
Given: a corpus $D$, text sequences $\vec{s_i} \in D$, input features (tokens) $f_k \in s_i$, a sequence encoder $E$, and hidden layer neurons $n_n \in E$, for each input token feature $f_k$ in the corpus sequences $s_i$, we calculate its: encoder neuron activations $\vec{a} = E(f_k)$; along with $\vec{a}$'s maximally active neuron $\vec{n_{argmax}} = argmax(\vec{a})$ and (maximum) activation value $a_{max} = max(\vec{a})$; to then record a \emph{single feature's activation} row vector $[f_k, n_{argmax}, a_{max}]$. If the encoder is part of a classifier model $C$, we also record the sequence's class probability $\hat{y} = C(s_i)$ and true class $y$ as a longer vector $[f_k, n_{argmax}, a_{max}, \hat{y}, y]$. For analyses in RQ1-3, we also record part-of-speech tags (POS, see \cref{sec:RQ1}) in the row vectors. This produces a matrix $M$ of neuron feature max activations that we aggregate to express each neuron as a probability distribution over maximally activated features in Step (2).

\textbf{(2) Preferred feature distribution per neuron:} From rows $m_r \in M$, we generate for each neuron $n_n$ its discrete feature activation distribution $A_{n_n}=\{(f_k, \mu(a_{max_1}, \ldots, a_{max_m}))|f_k, n_n, a_{max_j}\in m_r$ $\land \ m_r \in M \land n_{argmax} = n_n\}$, where each $f_k$ is a feature the neuron maximally activated on, and $\mu(a_{max_1}, \ldots, a_{max_m}) = \mu_{f_k}$ is the mean (maximum-)activation of that feature in $n_n$.
We then turn each activation distribution $A_{n_n}$ into a probability distribution $P_{n_n}$ by calculating the sum of its feature activation means $s_{\bar{\mu}} = sum(\mu_{f_1}, \ldots, \mu_{f_l})$ and dividing each $\mu_{f_k}$ by $s_{\bar{\mu}}$ to produce the normalized distribution $P_{n_n} = \{(f_1, \mu_{f_1}/s_{\bar{\mu}}), \ldots, (f_l, \mu_{f_l}/s_{\bar{\mu}})\} = \{(f_1, p_1), \ldots, (f_l, p_l)\} \}$, where, each $p_{f_k}$ is now the activation probability of a feature $f_k \in n_n$. Finally, for $n$ neurons in a model, $P$ describes their $n$ per-neuron activation distributions $P = \{P_{n_1}, \ldots ,P_{n_{n=|E|}}\}$.    

Features can be n-grams, and be tracked through multiple layers as in \cite{carter2019activation}. However, since in this work we focus on concisely presenting \method's transfer analysis, we only use uni-grams and a single layer.


\subsection{NEURON KNOWLEDGE CHANGE}\label{sec:hellinger}
We use \textbf{Hellinger distance} $H$ \citep{hellinger1909neue} and neuron distribution length $l$ to quantify differences between discrete feature preference probability distributions $p = P_{n_a}$ and $q = P_{n_b}$ of two neurons $n_a$ and $n_b$ as follows: 
\begin{align*}
    H(p, q) &= \frac{1}{\sqrt{2}} \sqrt{\sum_{f_k=1}^l (\sqrt{p_{f_k}}-\sqrt{q_{f_k}}})^2; \text{knowl. change} \\
    l(P_{n_n}) &= |\{f_k\ |\ f_k \in P_{n_n} \}|; \text{knowledge `diversity'}
\end{align*}
\textbf{Neuron length} $l$ describes the number of (unique) maximally activated features in a \activationhistorgram\ $P_{n_n}$. We use Hellinger distance because it is symmetric, unlike the Kullback-Leibler divergence. Importantly, if one of the preference distributions $P_{n_a}$ or $P_{n_b}$ is empty, i.e. has zero features (zero length), then the resulting Hellinger distance is ill-defined. Thus, Hellinger distance allows one to easily quantify neuron feature preference shifts to measure per-neuron knowledge change during pre-training (RQ1), zero-shot transfer (RQ2), and supervised fine-tuning (RQ3). 

Neuron length $l$ on the other hand allows us to define binary states like \textbf{`un-preferred'} for empty preference distributions ($l=0$) and non-empty ones \textbf{`preferred'} ($l>0$). We can use the two terms to classify \emph{three kinds of neuron preference state changes caused by different model training stages}: \textbf{`shared', `avoided', `gained'}. For `shared' neurons both distributions are non-empty (preferred) -- e.g. when neurons received maximum activations before and after retraining a model. `Avoided' neurons were active `preferred', but became less active `un-preferred' after retraining. Finally `gained' neurons, became more active after retraining, switching from `un-preferred' to `preferred' status. In RQ1-3 we will use changes in Hellinger Distance, distribution length and neuron states to identify which neurons overfit to few preferred features, which ones reuse features (transfer) and which one never specialize (unfit).

\section{EXPERIMENTS AND RESULTS}\label{sec:experimental}
We showcase \method's usefulness for analyzing and quantifying transfer in answering the previously stated research questions.
For RQ1, we pretrain an LSTM sequence encoder $E$\footnote{Though possible, we do not pretrain Transformers, due to high computation requirements, and since LSTMs encoders perform vastly better when pretraining on small collections -- compare \cite{transformerdiet} with \cite{DBLP:conf/iclr/MerityX0S17}. Instead, we focus on demonstrating \method's analytical versatility, especially for true low-resource scenarios, where large \pt\ is unavailable.} with $1500$ hidden units on \unsupervisedcorpus\ similarly to \citep{DBLP:conf/iclr/MerityX0S17,DBLP:conf/acl/RuderH18}, and apply (RQ2) or fine-tune it (RQ3) on \supervisedcorpus\ \citep{maas-EtAl:2011:ACL-HLT2011}, so we can analyze its zero-shot and supervised transfer properties. 
Each RQ's experimental setup and results are detailed below.

\subsection{RQ1: PRETRAINED WHAT KNOWLEDGE?}\label{sec:RQ1} 
In this experiment, we explore how pretraining builds knowledge abstractions. We first analyze neuron abstraction shift between early and later training epochs, and then verify that Hellinger distance and neuron length changes converge similar to measures like training loss.

\begin{figure}
\centering
\includegraphics[width=\linewidth,trim={0.21cm .21cm 0.3cm 0.21cm},clip]{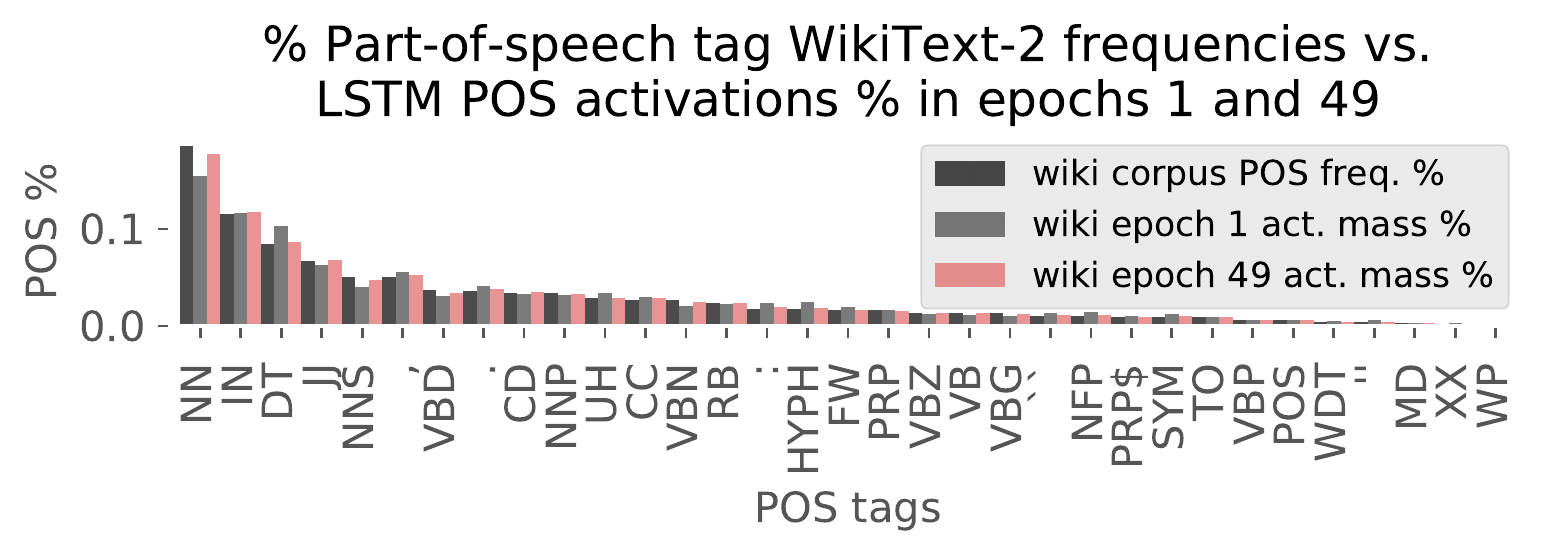}
\caption{\textbf{Encoder learns POS early on:} Black: Corpus tag frequencies (y-axis)  vs.\ encoder activations in \%-per-tag. Black: corpus frequencies via FLAIR. Grey: epoch 1 encoder. Red: fully trained encoder. POS is learned early, i.e. in epoch 1, confirming findings in \cite{LM_learn_POS_first}.}
\label{fig:POS_1_48}
\end{figure}

\begin{figure}
\centering
\includegraphics[width=\linewidth,trim={0.21cm 1.28cm 0.28cm 0.28cm}, clip]{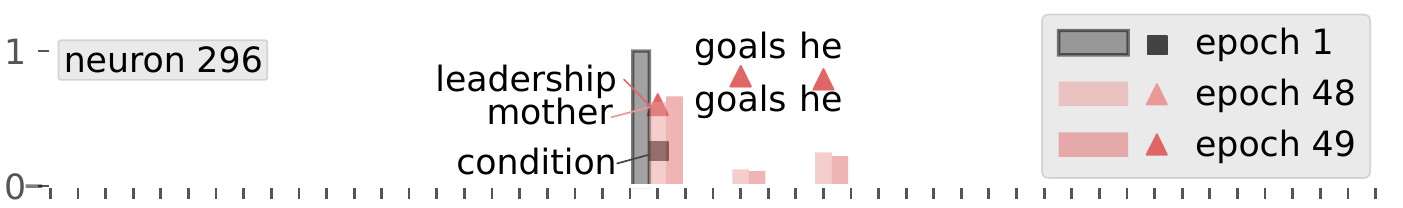}
\smallskip
\includegraphics[width=\linewidth,trim={0.21cm 0.3cm 0.28cm 0.42cm}, clip]{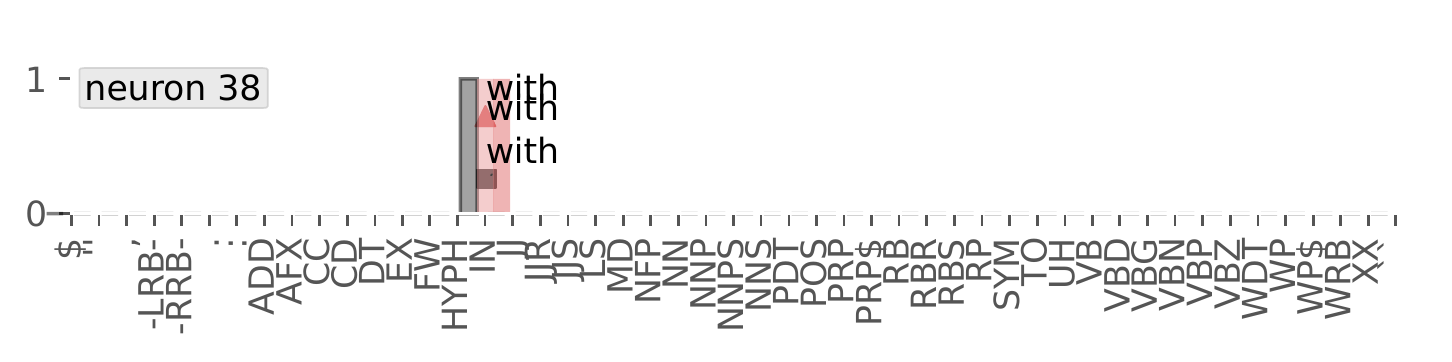}
\caption{\textbf{High vs.\ low Hellinger $H$ neurons:} Neuron token ($_\blacksquare$\textcolor{unsup_wiki_color}{$\blacktriangle$}\textcolor{t}{$\blacktriangle$}) and POS activation probabilities (bars) for epochs 1, 48, 49. Neurons with high $H$ (296) and low $H$ (38) between epochs 1:48.}
\label{fig:38_pt}
\end{figure}

We pretrain a single layer LSTM encoder $E$ on paragraphs from the \unsupervisedcorpus\ corpus $D_{wiki2}$ using a standard language modeling setup until loss an perplexity converge, resulting in 50 training epochs. We save model states at Epoch 1, 48 and 49 for later analysis. 
To produce neuron activation distributions $P_{wiki_{1}}$ (\colbox{unsup_1_wiki_color}{gray}), $P_{wiki_{48}}$ (\colbox{unsup_48_wiki_color}{pink}) and $P_{wiki_{49}}$ (\colbox{unsup_wiki_color}{red}) we feed the first 400.000 tokens of \unsupervisedcorpus\ into the Epoch 1, 48 and 49 model snapshots each to compare their neuron adaptation and incremental abstraction building using Hellinger distance and distribution length. Additionally, we record POS feature activation distributions using one POS tag per token, to later group tokens activations by their word function to better read, analyze and compare \activationhistorgrams\ -- see \cref{fig:POS_1_48}, \ref{fig:38_pt}, \ref{fig:ZS637} or \ref{fig:supervised_877}. POS tags are produced by the state-of-the-art Flair tagger \citep{akbik-etal-2019-flair} using the Penn Treebank II\footnote{\url{https://www.clips.uantwerpen.be/pages/mbsp-tags}} tag set.  

\begin{figure}
\centering
\includegraphics[width=\linewidth,trim={0.21cm .21cm 0.3cm 0.21cm},clip]{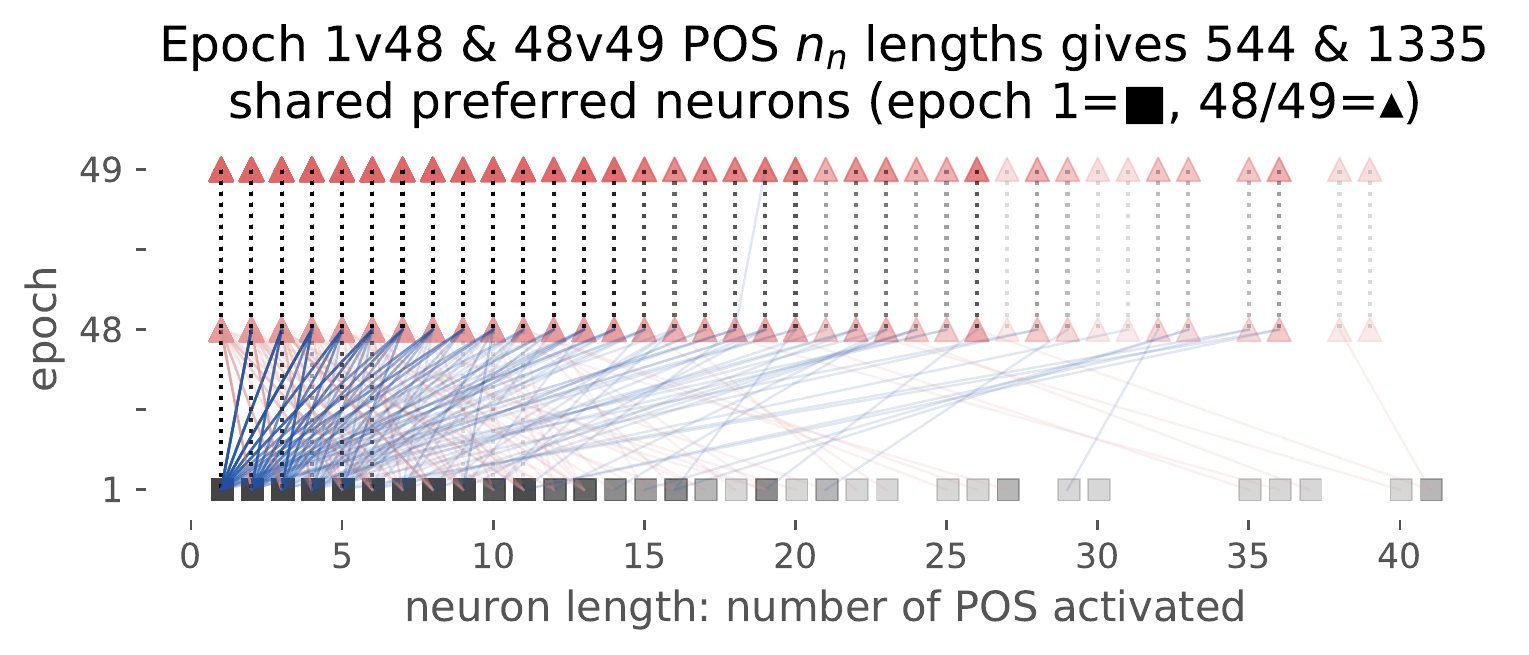}
\caption{\textbf{Pretraining neuron length shifts:} where neuron length $l$ (token variety) becomes; longer (blue $\slash$), shorter (red $\backslash$), unchanged (black $:$) for epoch 1, 48, 49. Token variety settles ($:$) in later epochs.}
\label{fig:length_shift_1_48_49}
\end{figure}

We use this experiment to verify the feasibility of using a \activationhistorgram\ approach, since comparing Epochs 1 vs.\ 48 should reveal \emph{large changes} to neuron abstractions, while Epoch 48 and 49 should cause \emph{few changes}. The resulting changes in terms of Hellinger distance, amount of `shared' preferred neurons, and \activationhistorgram\ lengths can be seen in \cref{fig:length_shift_1_48_49}.

While the Epoch 1 vs.\ 48 comparison produced 544 `shared' neurons, the later 48 vs.\ 49 comparison shows 1335 `shared' (\cref{sec:hellinger}) neurons. This means that pretraining the encoder distributes maximum input activations across increasingly many neurons. This can be seen in most neurons becoming longer (blue $_\blacksquare\slash$\textcolor{unsup_wiki_color}{$^\blacktriangle$} lines), and fewer neurons becoming shorter (red \textcolor{unsup_wiki_color}{$^\blacktriangle$}$\backslash_\blacksquare$ lines). As expected, for epochs 48 and 49 we see almost unchanged neuron length -- seen as dotted vertical (\textbf{:}) lines between epochs. Additionally, in later training stages, shorter neurons are more frequent than longer ones, reflected in the opacity of dotted vertical bars decreasing with neuron length. In fact, the average length of `shared' preferred neurons drops from 944.76 in epoch 1 to 524.55 and 519.34 in epochs 48 and 49. 

Since lengths of POS class preference distributions change significantly in the early epochs, we also analyze whether the encoders activations $P_{wiki_{1}}$, $P_{wiki_{49}}$ actually learned to represent the original POS tag frequency distribution of \unsupervisedcorpus. Thus, we express both corpus POS tag frequencies and encoder activation masses as proportional (relative) frequencies per token. In \cref{fig:POS_1_48}, we see relative corpus POS tag frequencies (black), compared with encoder POS activation percentages for epoch 1 (dark grey) and 49 (red). Evidently, the encoder learns a good approximation of the original distribution (black) even after just the first epoch (dark grey), which confirms findings by \cite{LM_learn_POS_first}, who showed that: ``language model pretraining learns POS first'', and that ``during later epochs (49) the encoder POS representation changes little''. Ultimately, the encoder near perfectly replicates the original POS distribution. We thus see that POS are well represented by the encoder, and that neuron adaptation and length shifts converge in later epochs in accordance with the quality of the POS match. This also tells us that \method, similar to more involved optimization-based analysis methods \citep{LM_learn_POS_first,SVCCA}, can reveal comparably deep insights into the mechanisms of \unsupervised\ training, while being simpler and more versatile (RQ1-3).

\begin{figure}
\centering
\includegraphics[width=\linewidth,trim={0.21cm .21cm 0.3cm 0.21cm},clip]{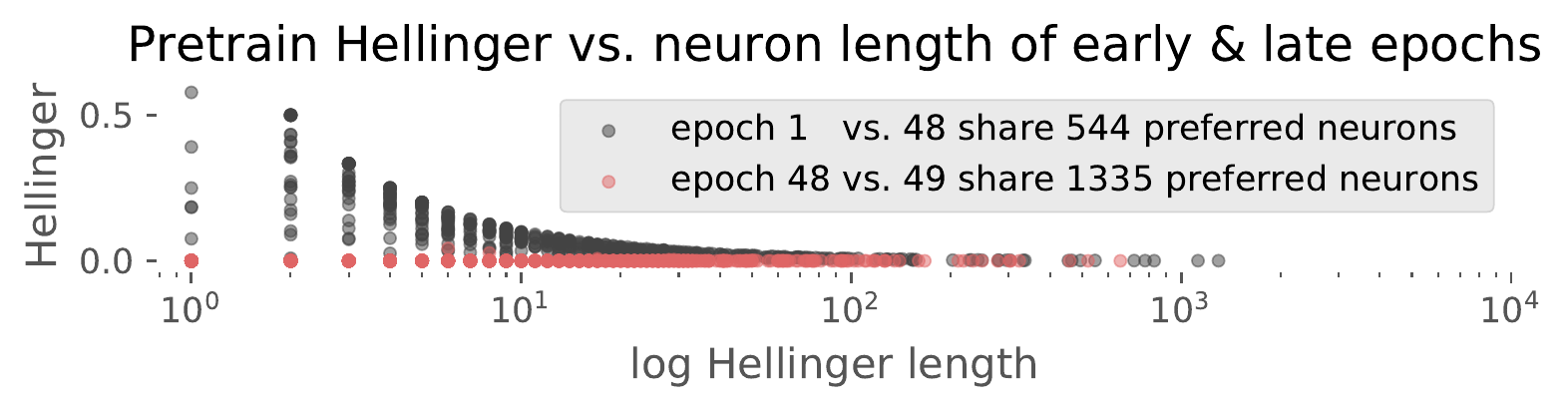}
\caption{\textbf{Encoder gains knowledge preference (neurons) through pretraining:} Later epochs activate more neurons maximally (544 to 1335), while Hellinger distance (knowledge change) reduces in later epochs (48 vs.\ 49, red line) vs. earlier epochs (1:48, black dots).}
\label{fig:length_hellinger_pretrain}
\end{figure}

Using \cref{fig:length_hellinger_pretrain}, a similar analysis about \emph{neuron feature distribution changes stabilizing at later training stages} can be made using Hellinger distances. When visualizing distances, we see that they shrink as expected by $99.92\%$ on average in later epochs and that neuron distance comparisons concentrate on medium length distributions of 10-200 features $f_k$ each. Preference distribution changes of short, specialized, neuron seem to produce higher Hellinger distances than longer, more general neurons. Since distances over different neuron lengths are not and should not be directly compared, this visualization acts to provide an \emph{explorable overview} of neuron distances over different preference distribution lengths, used to identify and examine interesting neurons in \emph{detail}.

To run such a detail analysis we pick 2 neurons from \cref{fig:length_hellinger_pretrain} for closer inspection of their \activationhistorgram\ changes between Epochs 1, 48 and 49. \cref{fig:38_pt} thus shows neuron \textbf{296} from the top 10 (head) most distant Epoch 1 vs.\ 48 neurons, and Neuron \textbf{38} from the 10 least changed ones (tail).
As expected from Neuron \textbf{296}'s high Hellinger distances between Epoch 1 and 48, we see that its token and POS distribution for Epoch 1, i.e., an outlined grey bar and the word `condition' ($_\blacksquare$), are very different from the Epoch 48 and 49 distributions (\textcolor{unsup_wiki_color}{$\blacktriangle$}, \textcolor{t}{$\blacktriangle$}), which show no significant change in token and POS distribution -- i.e.,\ they look nearly the same. Equally expected from Neuron \textbf{38}'s low Hellinger distance for Epoch 1 and 48; we see that it keeps the exact same token, `with', and POS, `IN`, across all three epochs. This demonstrates that Hellinger distance identifies neuron change, and that later epochs, as expected, lead to small neuron abstraction changes, while earlier ones, also as expected, experience larger changes.  


\subsection{RQ2: DO WE ZERO-SHOT TRANSFER?}\label{sec:ZS}
In this section, we analyze where and to what extent knowledge is zero-shot transferred when applying a pretrained encoder to text of a new domain -- without re-training the encoder to fit that new data. 

\begin{figure}
\centering
\includegraphics[width=\linewidth,trim={0.21cm .21cm 0.3cm 0.21cm},clip]{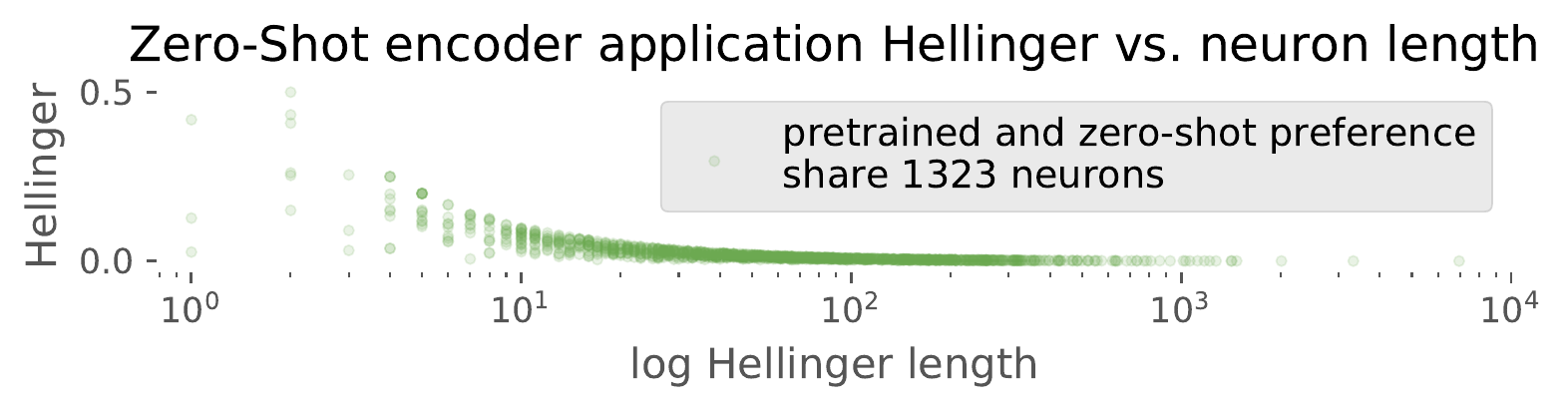}
\caption{\textbf{Neuron feature preference difference when applying to unseen text:} Hellinger distances between neuron \textbf{1323 `shared'} preference distributions $P_{wiki2}$ and $P_{imdb}$ on \unsupervisedcorpus\ and \supervisedcorpus.
}
\label{fig:ZS_hellinger_length}
\end{figure}

To do so, we apply the trained encoder $E$, in prediction-only mode, to both its original corpus \supervisedcorpus, $D_{imdb}$, and to the new domain \unsupervisedcorpus\ corpus $D_{wiki2}$, to generate \activationhistorgrams\ $P_{imdb}$ and $P_{wiki2}$ from the encoders' hidden layer, as before. We also record activation distributions for POS, which despite the FLAIR tagger being SOTA across several datasets and tasks, had noticeably low quality on the noisy \supervisedcorpus\ corpus. However on the \unsupervisedcorpus\ corpus, tagging produced comparatively sensible results. By comparing neuron token and tag activations $P_{imdb}$ (new domain) vs.\ $P_{wiki2}$  using Hellinger distances for the same neuron positions as in RQ1, we can now analyze zero-shot transfer as distribution shifts. Put differently, we estimate domain transfer between the pretrained model abstractions and text input from a new domain. High distances between the same neurons in $P_{imdb}$ and $P_{wiki2}$ tell us that the pretrained neuron did not abstract the new domain texts well, resulting in low transfer and poor cross-domain generalization.
When comparing $P_{imdb}$ and $P_{wiki2}$ in terms of Hellinger distances vs. neuron lengths in \cref{fig:ZS_hellinger_length}, we see that 1323 out of 1500 pretrained neurons ($88.2\%$) remain `preferred' (`shared') when applying $E$ to the \supervisedcorpus\ domain. A drop in the amount of `preferred' neurons compared to the RQ1 analysis, though at 1335 to 1323 small, is expected since the pretraining corpus covers a broader set of domains.

\begin{figure}
\centering
\includegraphics[width=\linewidth,trim={0.21cm 1.28cm 0.3cm 0.42cm},clip]{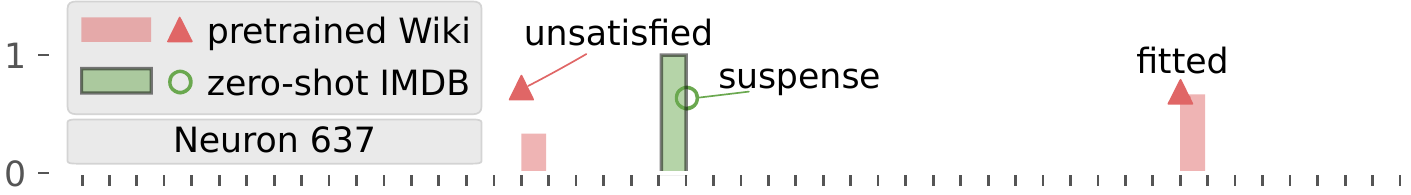}
\includegraphics[width=\linewidth,trim={0.21cm 0.3cm 0.3cm 0.3cm},clip]{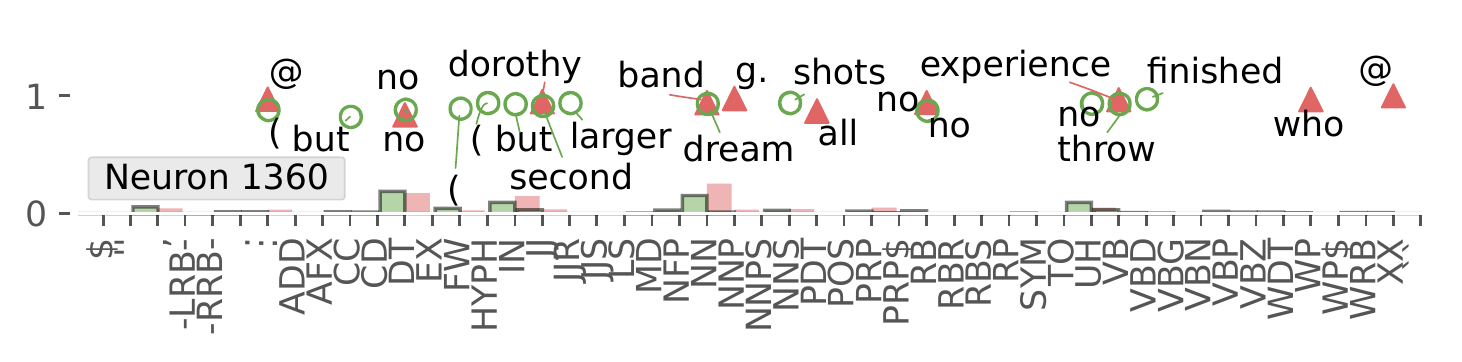}
\caption{\textbf{Low vs.\ high zero-shot transfer neurons:} Neuron \textbf{637} transferred little, while the `but-no' neuron \textbf{1360} transferred (applied) well from pretraining to the new \supervisedcorpus\ domain.}
\label{fig:ZS637}
\end{figure}

However, to gain a \emph{detailed} view of model abstraction behavior and zero-shot transfer, we analyze activation differences between $P_{imdb}$ (\colbox{unsup_imdb_color}{green}) and $P_{wiki2}$ (\colbox{unsup_wiki_color}{red}) for two specific neurons, visualizing one each from the 10 most (head) and 10 least (longtail) Hellinger-distant neurons. In \cref{fig:ZS637} (up), we see Neuron \textbf{637}, which has high Hellinger distance when comparing token feature distributions (\textcolor{t}{$\blacktriangle$}, \textcolor{unsup_imdb_dark_color}{$\circ$}). As expected, the neuron's feature preference between the pretraining corpus $P_{wiki2}$ and the new domain data $P_{imdb}$ changes a lot.
In fact, the distance in Neuron 637 is high in terms of both POS classes (word function semantics) and non-synonymous tokens -- see x-axis annotated with POS tags and tokens sorted by POS class. 
Overall, we see very \emph{little knowledge transfer} across data sets within Neuron 637 due to its \emph{feature over-specialization}, which is also observable in its short distribution length $l$ -- only 2 features activate. 
When looking at the low Hellinger distance Neuron \textbf{1360} in \cref{fig:ZS637} (lower plot), we see that the neuron focuses on tokens such as `no' on both datasets and `but' on \supervisedcorpus, suggesting that its pretrained sensitivity to disagreement (red), is useful when processing sentiment in the new domain dataset. Furthermore, we see that \supervisedcorpus\ specific tokens have many strong activations for movie terms like `dorothy' or `shots' (green). We thus conclude that Neuron 1360 is both able to apply (zero-shot transfer) its knowledge to the new domain, as expected from the low Hellinger distance, while also being adaptive to the new domain inputs, despite not being fine-tuned to do so, which is more surprising. In summary, we find that during zero-shot application of an encoder to new domain data, the pretrained encoder exhibits broad transfer, indicated by almost equal amounts of `shared' neurons between pretraining (1335) and application to the new domain data (1323). A supervision fit encoder however, has its knowledge reconfigured to superivsion, leading to much reduced transfer of pretrained knowledge, as we will see in RQ3. 


\subsection{RQ3: HOW DOES SUPERVISION BACK- TRANSFER LABEL KNOWLEDGE?}\label{sec:experiment_sup_transfer}
In this experiment, we analyze whether transfer constitutes more phenomena than just a high level observation like catastrophic \emph{forgetting}. Here, we want to see if knowledge also transfers `backwards' from supervised annotations to a pretrained encoder. Specifically, we analyze whether knowledge is added or discarded in two experiments. In Experiment 1, we demonstrate how \method\ can identify knowledge addition or loss induced by supervision at individual neuron level (\cref{sec:Supervision_adds_neurons}). In Experiment 2, we verify our understanding of neuron specialization and generalization by first pruning neurons that add or lose knowledge during supervision, and then measuring end-task performance changes (\cref{sec:Pruning_supervision_neurons}). Finally, we show how neuron activity increasingly sparsifies over RQ1-3 to gain overall insights about model-neuron specialization and generalization during unsupervised and supervised transfer (\cref{sec:Supervision_reduces_activity}).

\begin{figure}
\centering
    \includegraphics[width=\linewidth,trim={0.21cm .21cm 0.3cm 0.21cm},clip]{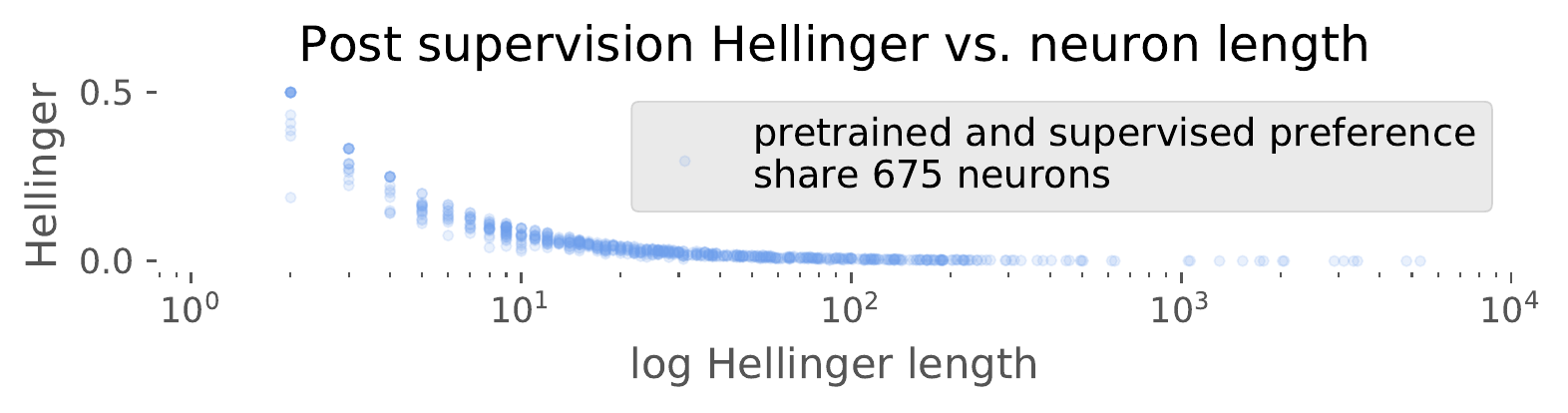}
    \caption{\textbf{Neuron feature preference change after supervision:} Hellinger distances of 675 `shared' neuron preferences before and after supervised encoder fine-tuning -- dropped from 1323.
    }
    \label{fig:Supervised_hellinger_length}
\end{figure}

For this RQ, we extend the pretrained encoder $E$ with a shallow, binary classifier\footnote{One fully connected layer with sigmoid activation that is fed by $E's$ end-of-sequence hidden state.} to classify \supervisedcorpus\ reviews as positive or negative while \emph{fine-tuning} $E$ to create a domain-adapted encoder $E_{\supimdbsuff}$. To guarantee a controlled experiment, we freeze the embedding layer weights and do not use a language modeling objective, such that model re-fitting is exclusively based on supervised feedback -- i.e.,\ on knowledge encoded into the labels. We tune the model to produce roughly $80\%$ $F_1$ on the \supervisedcorpus\ test set, to be able to analyze the effects of \emph{even moderate amounts of supervised fine-tuning before task (over-)fitting} occurs. To produce \activationhistorgrams\ $P_{\supimdbsuff}$, we feed the \supervisedcorpus\ corpus $D_{IMDB}$ to the newly fine-tuned encoder $E_{\supimdbsuff}$ -- i.e. using the same \supervisedcorpus\ text input. We also once more record POS tags for tokens. This time, since POS distributions are compared on the same corpus, their distances are more consistent than in RQ2. Analyzing Hellinger distance and neuron length change when comparing $P_{\supimdbsuff}$ vs.\ $P_{imdb-zero-shot}$ will tell us which neuron abstractions were changed the most \emph{due to supervision} -- i.e., show us `backward knowledge transfer'.
In \cref{fig:Supervised_hellinger_length}, we notice that only 675 neurons were `shared' compared to 1323 neurons in the zero-shot transfer setting (\cref{fig:ZS_hellinger_length}). In other words, \emph{supervision re-fits the sequence-encoder to `avoid' (unprefer) nearly half its neurons}. 

\subsubsection{Supervision adds and removes knowledge}\label{sec:Supervision_adds_neurons}
Somewhat surprisingly, supervision not only erased neurons, but also added distributions for 85 new neurons into $P_{\supimdbsuff}$ that had previously empty distributions in $P_{imdb-zero-shot}$. We analyzed these neurons and found that they represent new supervision task specific feature $f_k$ detectors. Below in \cref{tab:supervision_adds_neurons}, we show token features $f_k$ for the top three strongest firing neurons $n_n$ and the three least activating neurons out of the 85 -- i.e. supervision-specific neurons with the highest or lowest overall activation magnitude. Note: we removed stop-words like `the' or `a' as well as spelling duplicates from the table's feature lists to remain brief. Features are sorted by decreasing activation mass from left to right. We see that the first three highly active neurons roughly encode movie-related locations and entities as well as sentiment terms like `dull' or `great', though some seem unspecialized (general), fitting many genres.

\begin{table}
\caption{\textbf{Preferred features of 6/ 85 noisy supervision neurons gained by supervised fine-tuning:} 3 highly active ones (top 3), 3 seldomly active ones (bottom 3).}
\label{tab:supervision_adds_neurons}
\resizebox{\linewidth}{!}{%
\begin{tabular}{p{8.1cm}}
\hline \hline
\textbf{\#neuron : activation sum, features, (\#features total )} \\ \hline
200 : 1307.42 great, james, superb, famous, strange, possible, french, english, grand,  final, indian, solid \ldots\ (141) 
\\ \hline
1210 : 501.97 original, overall, good, real, some, dear, french, british, black, odd, italian, entire, many \ldots\ (161)\\ \hline 
125 : 299.12 more, two, best, one, few, most, three, nice, four, fellow, films, somewhat, lot, favorite, rare \ldots\ (77)\\ \hline \hline  
1289 : 7.92: terrific, dull, essential, celia, unbelievable, gentle, melancholy, intended, shaggy \ldots\ (14) \\ \hline 
372 : 4.18: walter \\ \hline
688 : 0.48: archer \\ \hline \hline
\end{tabular}%
}
\end{table}

When looking at the three least activating `supervision' neurons, we find more specialized feature lists. Some of them are short and very specialized to a specific feature -- e.g. the 372 `walter' neuron seems to be a `Breaking Bad' review detector, while `archer' (688) may detect the animated show of the same name. Somewhat surprisingly, Neuron 1289, despite only having a low activation sum, is comprised of many features that focus on sentiment like `terrific' or `dull', making the neuron more specialized than the top three. This suggests that `supervision' neurons with low activation mass, somewhat independent of their feature variety, are more specialized than the highly active ones -- which reflects in their lower `neuron length', i.e.\ them preferring fewer features. 
%
%
Detailed `discoveries' like supervision-gained knowledge reinforce our motivation, that an exploration-investigation approach can reveal detailed insights about a model's inner workings if `drilled-down'\footnote{A fundamental visualization techniques design pattern used to describe incrementally more focused analysis.} far enough, which underlines \method's application potential. 

\subsubsection{Pruning avoided, shared and gained neurons}\label{sec:Pruning_supervision_neurons}
To understand how much the `avoided', `shared' and 85 neurons `gained' by supervision affect predictive task performance, we run four pruning experiments (A-D) that remove neuron sets to measure the relative change from the unpruned $F_1$ score in $\%$ -- i.e., a drop from 80 to 77 is $\sfrac{77-80}{80}=-3.75\%$. Experiment (A) cuts 740 `avoided' neurons from the encoder $E_{\supimdbsuff}$, i.e., 740 neurons with empty \activationhistorgram\ after supervision. Experiments B and C cut the 20 least and most active neurons from the supervision tuned encoder. To select 20 neurons each, we sort neurons by their individual activation mass, i.e. the sum of a neuron's (max) activations, where `unpreferred' neurons with an empty preference distribution have zero activity. In the last pruning experiment (D), we prune the 85 neurons that became `preferred' after (due to) supervision -- i.e., were `unpreferred' before in $P_{imdb-zero-shot}$. \cref{tab:Pruning} shows for each pruning: the relative changes in training and test set $F_1$ and what percentage of the encoder's entire (max) activation mass the pruned neurons drop. 
\begin{table}
\caption{\textbf{Pruning avoided, preferred and supervision-gained neurons:} After supervised encoder fitting; (A) prunes avoided (unpreferred) neurons, (B,C) prune the least and most preferred neurons, and (D) prunes 85 neurons gained by supervision -- i.e., that were non-preferred in pretraining. Colors represent relative score change in $\%$ from original -- score drops (red,$-$), gains (blue). Similar to \cite{Lottery1}, test score increases, despite pruning $\approx 50\%$ of encoder neurons in (A).}\label{tab:Pruning}
\centering
\includegraphics[width=\linewidth]{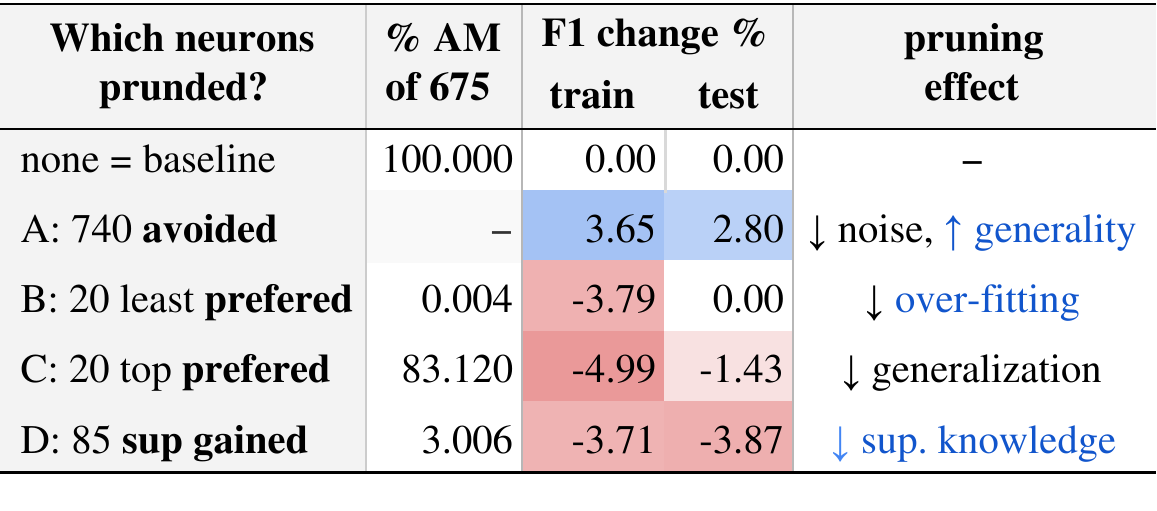}
\end{table}

For pruning experiment \textbf{(A)}, we see that removing `avoided' neurons not only does not drop performance as commonly observed when dropping irrelevant neurons \citep{transformerheads19,MovementPruning}, but actually \emph{increases both training and test set} performance by 3.65 and 2.80 $\%$ respectively, resulting in better generalization.
In Experiment \textbf{(B)}, when removing seldomly activated supervision neurons, as indicated by the low activation mass percentage of $0.004\%$, we lose significant training performance ($-3.79\%$), but no test set performance, telling us that those neurons were over-specialized or over-fit to the training set. It also tells us that these neurons were likely short (over-specialized), similar to those in \cref{tab:supervision_adds_neurons} that have low activation mass (372, 688). When we examined this intuition, we found that each of the 20 neurons has a length of exactly one -- i.e. is over-specialized.  
When pruning the 20 most heavily used supervision neurons \textbf{(C)} with $83.12\%$ (max) activation mass, we see the largest drop in training set performance out of all experiments (A-D). This tells us that, similar to observations in experiment (B), \method\ again identified neurons that strongly over-fit to the training data, while they overfit the test set to a lesser extend.
Thus, Experiments (B, C) indicate that cutting supervision specific neurons after training can help preserve generalization performance, i.e.,\ reduce generalization loss. 
Lastly, for \textbf{(D)}, when pruning the 85 neurons `gained' by supervision both training and test performances drop by equal amounts. Since these 85 supervision-only neurons only became `preferred` after supervised fine-tuning, this indicates that pretraining-exposed neurons as in (B) and (C), suffer less from overfitting on new (test set) data, even when pruned. We reason that pretraining-exposed neurons in (B) and (C) have their knowledge partially duplicated across other neurons, while the supervision-only knowledge in the 85 `gained' neurons (D) has no such backups. 
\textbf{(Neuron) generalization, specialization:} These observations are not only consistent with known effects of pretraining on generalization \citep{TuneNotTune19,DBLP:conf/acl/RuderH18}, but also show that \method\ can identify and distinguish at \emph{individual neuron level}, which parts of a neural network improve or preserve generalization (A, B) and which do not (C, D). Moreover, the pruning based generalization increase in experiment (A) is consistent with findings of Lottery Ticket based pruning by \cite{Lottery1}, as well as with our notions of \emph{neuron specialization an generalization} used throughout \method. This demonstrates the method's effectiveness in identifying neurons that affect generalization and specialization.   

\begin{figure}
\centering
\includegraphics[width=\linewidth,trim={0.21cm 1.28cm 0.3cm 0.42cm},clip]{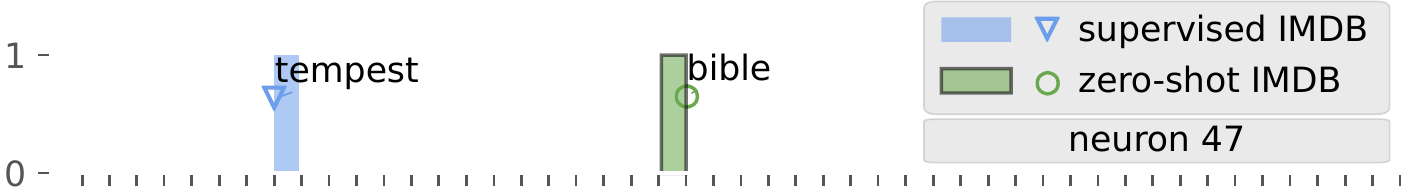} 
\includegraphics[width=\linewidth,trim={0.21cm 0.3cm 0.3cm 0.1cm},clip]{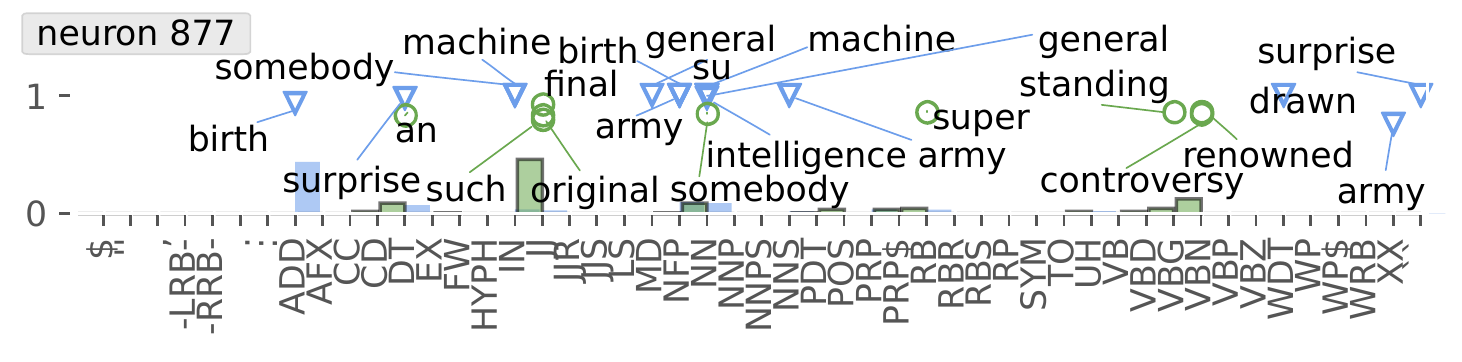} 
\caption{\textbf{Low and transfer to supervision:} Neuron \textbf{47} saw no transfer, while Neuron \textbf{877} transferred its knowledge better from before (zero-shot) to after supervised fine-tuning on \supervisedcorpus.}
\label{fig:supervised_877}
\end{figure}

To again analyze what individual neurons learned, we inspect neurons with high and low Hellinger distances between encoder activations before (\colbox{unsup_imdb_color}{green}) $P_{imdb-zero-shot}$ and after supervision (\colbox{sup_imdb_color}{blue}) $P_{\supimdbsuff}$. In  \cref{fig:supervised_877}, we show Neuron \textbf{47} (up), from the top 10 highest Hellinger distances. 
We see that the neuron 47 changed in both POS and token distributions after supervision, which suggests catastrophic forgetting, or supervised reconfiguration. 
For the low Hellinger distance Neuron \textbf{877} (down), we see some POS and token distribution overlap before and after supervision, and that movie review related terms (green \textcolor{sup_imdb_color}{$\triangledown$}) become relevant, compared to noticeably war related tokens before supervision (green \textcolor{unsup_imdb_dark_color}{$\circ$}). This shows the neuron's semantic shift (POS, token) due to supervision -- i.e., limited knowledge transfer occurred despite the low Hellinger distance.
Moreover, distribution length changed for this neuron from 9 before to 15 tokens after supervision, indicating a lack of transfer. Finally, we recall that in the zero-shot case more neurons were `shared' than after supervision, 1323 vs.\ 675 (\cref{fig:ZS_hellinger_length} vs.\ \cref{fig:Supervised_hellinger_length}), which should be reflected in the overall activation magnitude produced by encoder $E$ before and after supervision.

\subsubsection{Supervision sparsifies neuron knowledge}\label{sec:Supervision_reduces_activity}
\begin{figure}
\includegraphics[width=\linewidth,trim={0.21cm .cm 0.9cm 0.cm},clip]{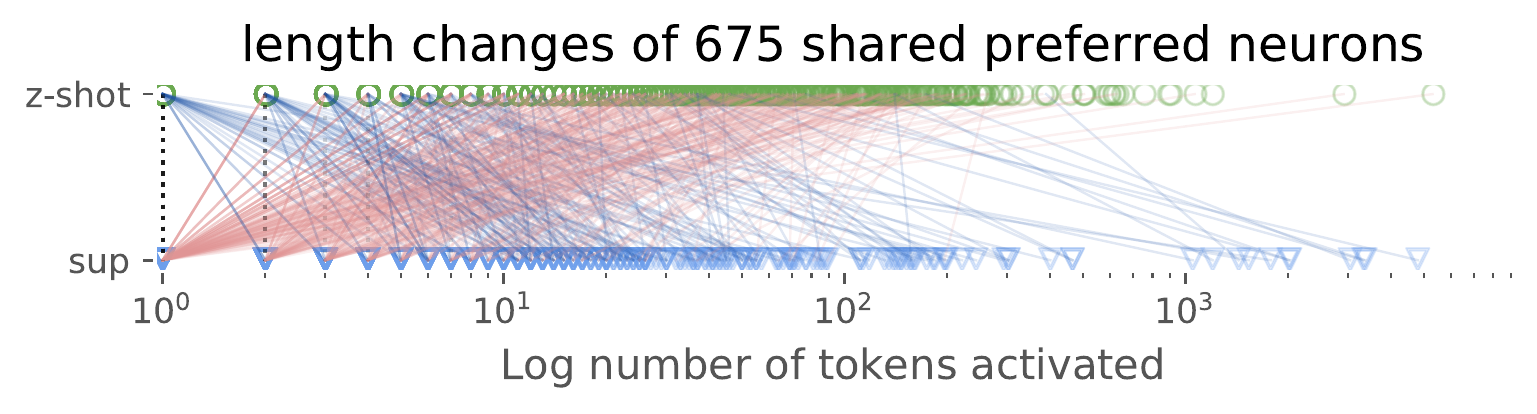}
\caption{Neuron length before ($\circ$) and after \textbf{sup}ervision ($\mathbf{\triangledown}$). After supervision $\slash$ neurons are shorter, $\backslash$ are longer, and $:$ are unchanged.}
\label{fig:supervised_length_shift}
\end{figure}

\begin{figure}
\includegraphics[width=\linewidth,trim={0.21cm .21cm 0.3cm 0.21cm},clip]{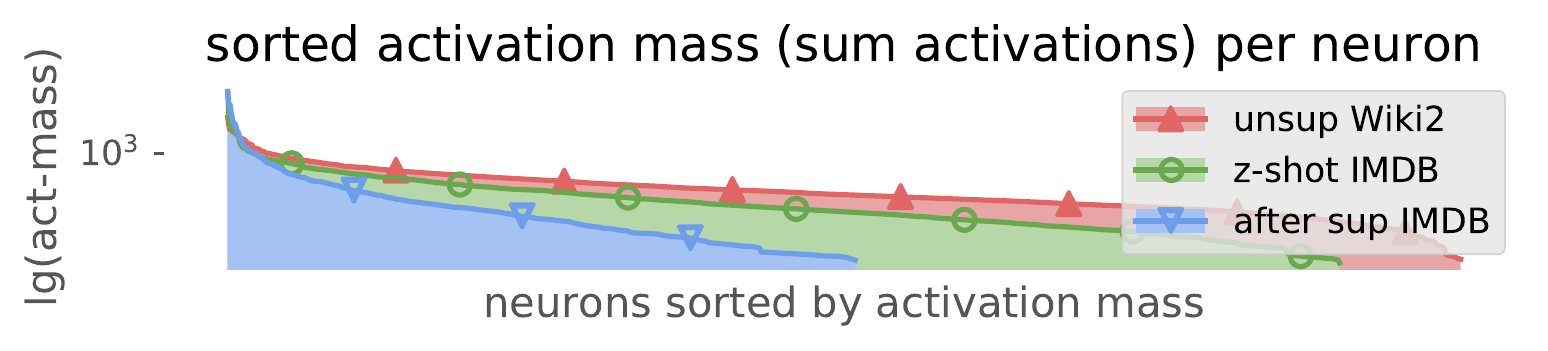}
\caption{Sorted neuron activation masses, for the pretrained (large, \st{$\blacktriangle$}), zero-shot (middle, \st{$\circ$}), and supervision tuned encoder (small, \st{$\triangledown$}). Supervision sparsifies activations -- i.e. \st{$\triangledown$} head peaks, tail shortens.}
\label{fig:activation_sum}
\end{figure}

To investigate the distribution length shift and activation sum hypotheses formulated above, we visualize the shift of neuron length before and after supervision (\cref{fig:supervised_length_shift} and \cref{fig:activation_sum}), as well as the activation mass for the three research questions: (RQ1) pretraining, (RQ2) zero-shot, and (RQ3) supervision.

In \cref{fig:supervised_length_shift}, we see neurons that shortened (red lines, $_\triangledown\slash^\circ$), or got longer (blue lines, $^\circ\backslash_\triangledown$), after supervision. Token preference distributions of neurons actually \emph{slightly lengthen} by $4.62\%$ on average over the 675 shared neurons,\footnote{Over the entire 1500 neurons, neuron token length shortens by $42.53\%$ after supervision.} while POS preference distributions, severely shorten at $32.83\%$ (not shown). Similar neuron lengthening, `feature variety increase', from supervision, was already apparent in neuron 877 (\cref{fig:supervised_877}), where supervision appeared to have specialized and extended a previously unspecific neuron into a movie sentiment detector\footnote{Again, without deeper analysis, we are not claiming that this is the case, only that such points for investigation and new, interesting hypotheses can be identified via \method.}.  

In \cref{fig:activation_sum}, we see that the activation mass -- i.e., the sum of activation values -- differs across corpora and encoder activation distributions $P_{imdb-zero-shot}$, $P_{\supimdbsuff}$ and $P_{wiki2}$. A much more peaked activation mass is produced after the encoder has been fine-tuned via supervision and then again applied to \supervisedcorpus\ (blue, \st{$\triangledown$}) compared to before supervision (green), which is a strong indicator that supervision sparsified the neuron activation and therefore the abstractions in the encoder.
The activation mass of the pretrained encoder $E$ on its pretraining corpus (\unsupervisedcorpus, red \st{$\blacktriangle$}) is, unsurprisingly, the broadest, while it activates less strongly on the same amount of text (400k tokens) on the \supervisedcorpus\ text (green, \st{$\circ$}), due to the mismatch of domains between pretrained encoder and the new data domain -- as seen in RQ2.

\section{RELATED WORK}
Recent explainability methods \citep{CSI19,belinkov-glass-2019-analysis,ExplainingExplanations,Atanasova2020ACL} fall into two categories: \emph{supervised} `model-understanding (MU)' and `decision-understanding (DU)'. DU treats models as black boxes by visualizing how important each input is for a prediction outcome to understand model decisions. MU enables a grey-box view by visualizing internal model abstractions to understand what knowledge a model learned. Both DU and MU heavily focus on analyzing supervised models, while understanding transfer learning in self- and supervised models remain open challenges. \textbf{Supervised `DU':} techniques explain decisions for supervised (probing) tasks to \emph{hypothesis test} models for language properties like syntax and semantics \citep{Senteval,LRP_graphs}, or language understanding \citep{Glue,DiagnosticClassifiers}. DU is limited to \emph{supervised} analysis of \emph{individual} samples \citep{ExplainingExplanations,arrasACL19}.
\textbf{MU:} techniques like Activation Atlas or Summit \citep{carter2019activation,hohman2019summit} explore supervised model knowledge in vision, while NLP methods like Seq2Seq-Vis \citep{SEQ2SEQVIS} compare model behavior using many \persample\ explanations. However, these methods produce a high cognitive load, showing many details, which makes it harder to understand overarching learning phenomena. 
\textbf{(Un-) supervised `model and transfer understanding':} \method\ modifies ideas behind activation maximization \citep{ActivationMaximization,olah2017feature,carter2019activation} (see \cref{sec:approach}) to enable measuring neuron knowledge change, specialization and generalization as well as to guide explorative transfer analysis by \emph{quantifying interesting starting points}. Somewhat similarly to our setup in RQ3, \cite{singh2019medial} ``calculate Helliger distances over `neuron feature dictionaries` to measure neuron adaptation during `supervised' task learning'' in the prefrontal cortex of rats. Measuring changes in neuron \activationhistorgrams\ enables fine-grained analysis of neuron (de-)specialization and model knowledge transfer in RQ1-3. \method\ extends upon probing task and correlation based transfer analysis methods like \cite{CR_transfer, Important_neurons, SVCCA}, to provide more flexible, yet nuanced, (un-)supervised transfer \interpretability\ and  analysis for current and future (continual) pretraining methods \citep{TuneNotTune19,RuderEpisodic2019}, while also enabling \emph{discovery of unforeseen hypotheses} to help \emph{scale learning analysis beyond the limitations of supervised probing and approximate correlation analysis}.

\section{CONCLUSION AND FUTURE WORK}
We presented \method, a simple, yet nuanced model knowledge explainability method for analyzing how neuron knowledge transfers between pretraining (RQ1), zero-shot knowledge application (RQ2), and supervised fine-tuning (RQ3). We showed how to extract neuron knowledge abstractions in NLP, developed extensible explainability visualizations and demonstrated how this can measure knowledge abstraction change. 
We find that \method\ enables explorative analysis of how knowledge is lost and added during supervision (RQ3), how neurons overfit or generalize (RQ1-3), and how pretraining builds knowledge abstractions (RQ1). \method\ is designed to reduce computational and cognitive load, but is flexible and scalable. In future, we will use \method\ for more advanced transfer models and metrics. The code and visualizations are available at \href{https://github.com/copenlu/tx-ray}{github.com/copenlu/tx-ray}. 

\subsubsection*{Acknowledgements}
This work was supported by the German Federal Ministry of Education and Research within the projects XAINES and  DEEPLEE (01IW17001). 

\bibliography{arxiv.bib}
\bibliographystyle{apalike}
\end{document}